\documentclass[runningheads]{llncs}
\usepackage{graphicx}
\usepackage{hyperref} 
\usepackage{url}
\begin{document}
\title{Computer Vision for Supporting Image Search}

\subtitle{(A  Compelling Application for Vision Processing)}

%
%
\author{Alan F. Smeaton\orcidID{0000-0003-1028-8389} }
\authorrunning{A.F. Smeaton}
%
\institute{Insight Centre for Data Analytics,\\ Dublin City University, Glasnevin, Dublin 9, Ireland\\
\email{alan.smeaton@dcu.ie}
}

\authorrunning{A.F. Smeaton}
\titlerunning{Computer vision for supporting image search}
\maketitle              
\begin{abstract}
Computer vision and multimedia information processing have made extreme progress within the last decade and many tasks can be done with a level of accuracy as if done by humans, or better. This is because we leverage the benefits of  huge amounts of data available for training, we have enormous computer processing available and we have seen the evolution of machine learning as a suite of techniques to process data and  deliver accurate vision-based systems.
What kind of applications do we use this processing for ? We use this in autonomous vehicle navigation or in  security applications, searching CCTV for example, and in medical image analysis for healthcare diagnostics. 
One application which is not widespread is image or video search directly by users.
In this paper we present the need for such image finding or re-finding by examining human memory and when it fails, thus motivating the need for a different approach to image search which is outlined, along with the requirements of computer vision to support it.

\end{abstract}
\section{Introduction}

Digital forms of multimedia now assault our senses and is embedded into our lives for work, leisure and entertainment.  Technology support for creating, capturing, compressing, storing, transmitting and rendering multimedia are now mostly solved while the technology we use for finding multimedia information, or for having it find us, has not developed as much.

We use computer vision techniques to analyse visual imagery and across a range of applications, computer vision supports those applications well.  Systems for medical diagnosis based on imaging, for autonomous driving, for event detection and classification in surveillance video, for highlights detection and video summarisation in broadcast media, and now even image and video generation and style transfer all use different forms of computer vision

In this paper we are particularly concerned with visual search, searching for images and videos, for which there are many use cases. 
Finding, or re-finding images is a niche scenario in which we want to locate or re-locate an image or video clip we think we may have seen in the past.  This could be an image on a web page, or in an advertisement, or in a movie, or on social media and we may be able to recall certain aspects of it but not describe it fully. The challenges presented here are much more complex than re-finding  an equivalent text document because when re-finding text we will generally recall  the  words or phrases used and will have enough recollection to adequately describe the document's content. When re-finding imagery we will usually have only a partial recollection of the image and thus we cannot describe it fully.
Furthermore, we know that since as much as 40\% of searching is re-finding information \cite{10.1145/1718487.1718512} then an even greater \% of image search must be  re-finding.

When we examine the kind of computer vision used to support  visual information seeking we see a mis-match and in this paper we  re-examine the relationship between searching to find or re-find visual content, and the computer vision used to support it.
Since so much of our visual information seeking is re-finding, human memory is core to this so the paper will also include an overview of human memory and why we forget and thus have to re-find information. We will see  how we are wrapping ourselves around the existing search technology behind computer vision and not how the technology is wrapping itself around us, to support us.
This “technology first, human second” approach is bad because it makes us change  or to develop technology for visual search systems in this case, that are less than they could be and instead of supporting us, we are forced to change

In the next section of this paper we provide a recap on human memory and how it works and why it sometimes fails, followed by an overview of information seeing and how search is just one part, the final part. We then look at  search systems and how they have developed and then we examine  computer vision as it is used in visual information seeking. Finally in the conclusions section we argue for visual  search as a conversation between searcher and repository, and what form of computer vision is needed for this.

\section{Human Memory}

The human brain is  made of approximately 86 billion neurons, known as grey matter, connected by trillions of connections (synapses) along which electric pulses are transmitted.
This architecture of a huge number of simple connected processors, is now realised as being very good for solving very  complex problems, like vision, and learning.
Human memory, a core component of all brain processing, is a brain-wide process and is broadly divided into 3 types:
\begin{enumerate}
\item Sensory or short term memory where we see, hear or feel something and which has a rapid decay unless it is refreshed;
\item Sensor memories are then passed to our
short term memory but retained only if we are attentive to those senses. Usually we can only retain up to 7 items or chunks of information in our short term memory, and these are then discarded unless we consciously decide to store them in our long term memory;
\item Long term memory is a more permanent form of memory and is made of (1) declarative memory which is facts and events, like the answers to quiz questions and (2) procedural memory which is like muscle memory, playing piano or performing some work-related tasks like composing and sending an email.
\end{enumerate}

There are many commonly-observed  memory phenomena like false memories in autobiographical recall,  Proustian moments of recall where sensing something, like the smell of garlic or hearing a certain song triggers a cascade of recollections about something else, or flashbulb memories where we have vivid recollection around significant events in our lives, like where we were and what we were doing when we heard the news about a significant world event or a personal event like the bereavement of a close friend.

Throughout our lifetime, our memory fails, naturally.  Sometimes it can fail from a mild degree to a severe degree, and this is always from neurological damage.
This can be immediate damage such as with a trauma, or it can be progressive such as with age. 
It can be caused by drugs like alcohol, by dementias like Alzheimer’s, by diseases like Huntington’s or Parkinson’s, by OCD, schizophrenia, stroke, or Tourette's syndrome. 
But even without any of these factors, we just forget things anyway during our lives at work and at leisure, and in our social interactions with others.  This is because not everything moves from our short term memory to our long term memory and even within our long term memory our memories will decay.

Investigations into how long we remember were first carried out by Hermann Ebbinghaus in the 1880s \cite{ebb1880}, and the results of his work still hold true, that there is an exponential loss from our long term memory unless information is reinforced \cite{murre2015replication}.
That means that last-minute cramming before an examination usually won’t work and it means that the best way is to learn something, forget some of it but learn it all again, forget some again but less than before, then learn it all again, etc.

How steep is this forgetting curve and how long are we likely to remember things~?
In the time it takes to make and drink a cup of tea we forget up to 42\% of what we learned, 
in an 8-hour work day we forget 64\%, and in a week it will be 75\% of what we learned and wanted to remember, that will be forgotten.

To stop memories from decaying and fading, we need to rehearse and to repeat, and to revise through spaced repetition.  
What is attractive about the forgetting curve is the steepness of the tail-off. The first time something new is learned it is quickly forgotten, the second time is less quickly forgotten and as you review and review, what’s learned remains remembered for longer.
So repetition –- especially spaced repetition -– helps us remember more, and for longer.

There are  memory ``tricks” we can to re-enforce our memory like connecting something new we want to remember with existing information we recall, thus weaving it into things we already know.
This is achieved using things like mnemonics, word association, visualisation, chunking or breaking complex information or that big picture into smaller bits.
The internet and popular science is full of resources which claim to improve your memory but they are basically exploiting these two tricks \ldots  repetition to overcome the forgetting curve, and the weaving of new information into your existing knowledge.

We would like to believe that technology, in all its forms, is developed to help people.
We have, and we use, technologies to help our sight (glasses, contact lens, smartphones which automatically caption our surroundings), our hearing (cochlear implants), and our movement (bionic prosthetics).  This is quite accepted because we have used technology for some of our other basic needs like transport and communications.

Human memory already benefits  from the use of supporting technology. A trivial example is phone numbers, which we now don't have to remember when we call somebody because they are replaced with aliases, the destination person's name, on our phones.
We have notepads and post-its and chunking and tracking devices for our keys, calendars and diaries and more, but we have not planned a technology from scratch to help our memories, we have evolved it over time, especially in recent times and almost by stealth.   Because so much of our lives are spent online and digital, we have learned that we do not have to remember things to the same degree as previously because we can always search for things from our past. This includes our emails, our documents, our browsing history, our enterprise systems and our social media feeds.

So before we take a  look at how we find things among these enormous, up-to-date, always available information resources, and see how easily accessible they really are, let us take a look at information seeking more generally.

\section{Searching and Information Seeking}

Searching for information or finding information is known in the research world as ``information retrieval'' (IR) and as a scientific topic within the broad computing area it is not  new .  IR pre-dates web search engines and was not a research backwater but  a slowly maturing discipline. It takes time for the science behind a research area like information science to develop, it  cannot be pushed or forced and by the 1990s we had evolved to a model of search based on a ranked list of search outputs. This ranked list was based on the statistical distributions of words in documents and in queries. We were developing interaction models where searches were incremental, looking beyond simple search ranking. We were looking at aspects like the incremental value of additional information in a ranking and relevance feedback in mid-search so as to re-rank as yet unseen documents. Though these were  computationally demanding they would support searching as an interaction, a journey between the searcher and the information repository with the final destination being a user with a satisfied information need.  

Throughout the intervening years IR  developed to include things like the vector space model, term weighting approaches like BM25 and then language modeling, all built around the frequency of word occurrences in text and document representation based on a bag-of-words or  more correctly a list of words.  More recently we have seen retrieval based on  learning-to-rank approaches, leveraging logged data from previous searches and advancements in machine learning.

\begin{figure}[!htb]
\centering 
\includegraphics[width=0.6\textwidth]{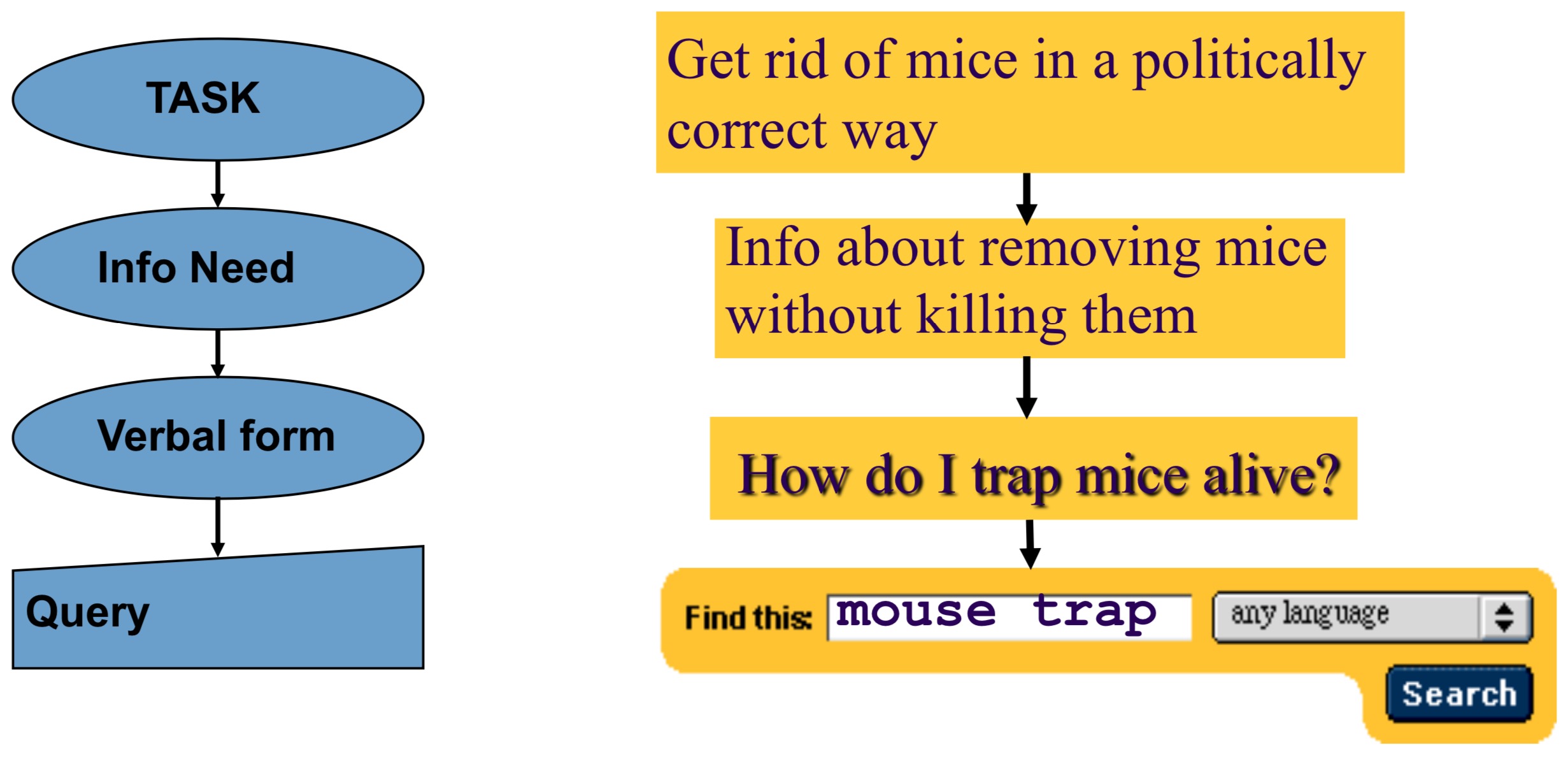}
\caption{How we go from information need, to search query (taken from (\cite{mannbook}).} \label{fig1:infoneed}
\end{figure}

Figure~\ref{fig1:infoneed} taken from  \cite{mannbook} illustrates the relationship between tasks we are performing, for work or for leisure, how these tasks generate information needs which we verbalise as a query which is then used as input to a search engine.  What should be clear  is that there is a large gap between what our task wants us to do, and the words we type into a search system, a kind of semantic gap. 

Then in the mid-1990s came the web, and the need or opportunity to provide search across all of the web and this was realised using a model of stateless search with no continuity or supporting the information seeking goal that enveloped each individual search. Instead the focus was on  really fast response time to re-enforce the illusion that you were getting somewhere fast with your search.  That was, and remains, the business model of fast response time and targeted advertising, and we accepted it because it was fast and it was free. Very quickly  we grew used to this as being the way that search operated and it was superb engineering because it delivered fast response time in searching up-to-the-minute repositories of information.

In the intervening years the effort in terms of deployed web and enterprise searching has been towards crawling and indexing, and the search function has developed things like a multitude of positive and negative ranking factors for web pages and for websites and link importance and PageRank, and all of the ``smarts'' like Panda/Farmer and Penguin, but the search remains the same \ldots fast response time but stateless search.  There have been some   exceptions like the knowledge graph, introduced by Google in 2012, which presented facts extracted from pages instead of ranking.

As a service, search is free, even though we pay for it with our data, but as a quality service it is not very good. That is because we have grown accustomed to searching poorly because the incremental cost to us of doing another search is so little, precisely because it is so fast. Web search and enterprise search use a brute force approach and our over-arching tasks, the things we actually want to find in terms of our information seeking, information finding, information discovery, these are  a mis-match to the rapid-fire, stateless searching on the web and on our enterprises.  There is no support for negotiated discovery, no information exploration, no information seeking, and as users we do not miss this  because we never had it. 

Information retrieval is hard, and it is difficult not because of the information we search for but because of us, the people who do the searching. We are all different and our diversity is what makes humankind.  That means that building a search system with a one-size-fits-all and no tailoring for the individual, is already a flawed choice.

When we want to find information, we need it in order to make decisions, and that might be tasks related to  work, leisure, or social interactions during interactions  with others.  When we look at these information needs we discover that there are different kinds of information need.
Our information needs may be 
verificative or explorative so for example {\em ``who is the President of the European Commission''} vs. {\em ``tell me about the European Commission''}.  Our needs might be 
precise or vague
such as {\em ``Where was the movie Castaway shot ?''} vs. {\em ``tell me about the movie Castaway''}. We may be interested in topics over a
long lived or a short duration and finally our needs might
cost-dependent or inconsequential with respect to cost, where cost is measured as our own time.

\begin{figure}[!htb]
\centering 
\includegraphics[width=\textwidth]{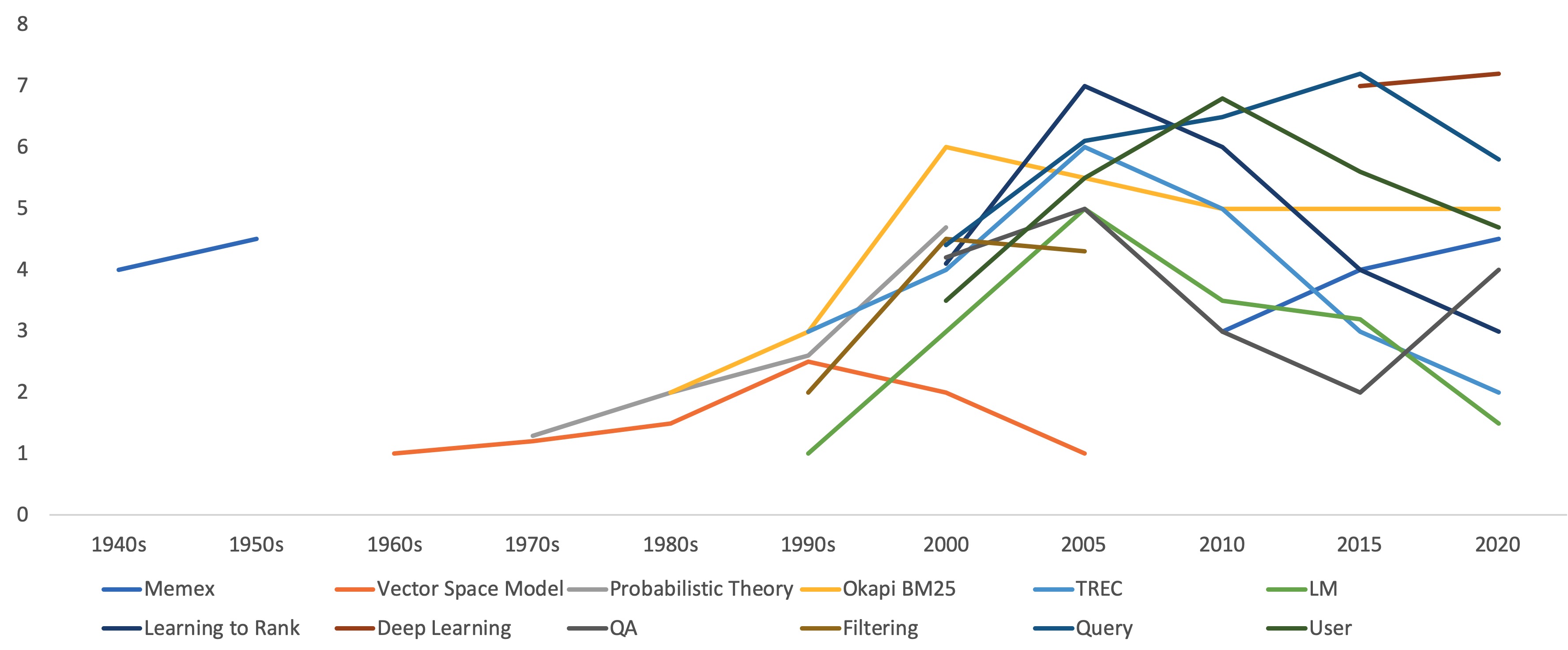}
\caption{History of the major research topics in information retrieval (taken from (\cite{yang2019}).} \label{fig1:hist}
\end{figure}

This interpretation of information seeking is not new and  information scientists  have been developing theories of information seeking for decades. This is done in an attempt to understand the processes behind information seeking so we can then build tools to support different stages.   Often this is done within narrowly defined tasks, like information seeking to support journalists writing articles, or in seeking information about health.  Figure~\ref{fig1:hist} taken from \cite{yang2019} shows  research areas that have come and gone over the last several decades as major foci for IR research and most of these still subscribe to this single query, stateless, ranked output model of searching rather than addressing information seeking. 

Meanwhile the rate at which our information access is evolving is so fast that the models of information seeking cannot keep up with the technology. Think of things we use now, every day, which were not present a decade ago, or even just a couple of years ago.  The Apple iPhone was introduced in 2007 and iPad in 2010, 
Amazon Echo was introduced in 2017, Twitter started in 2006, Snapchat in 2011 and TikTok was founded in 
2016. The reason for mentioning these is to show how the information seeking landscape, the ways we can find information or it find us, the kinds of information, delivery platforms, all this  changes hugely, so modelling information seeking is really difficult as a result.

As yet another further complication, nowadays we all have continuous partial attention as we work and relax. Many of us now do a kind of multi-tasking because our devices -- smartphones, tablets and laptops -- are multi-purpose.   Sometimes this happens when we mindwander which we all do, all the time, skimming the surface of incoming information as it is presented to us, picking out  relevant details and moving on to the next stream in a continuous rather than  episodic fashion, casting a wide net across our tasks, but not often giving any of them our full attention.

If we accept that there is a ``semantic gap'' between our information seeking and the search tools that we use, what does search look like, and we answer that in the next section.

\section{Search Systems}

We know that search is not easy,  it is indefinite, with lots of uncertainty and with complex inter-dependencies, user requirements are incomplete, sometimes contradictory and always changing as searchers' needs evolve, even in mid-search.
The standard search interface and affordance  allows for the very simple concepts of a search
query as a set of words or query terms, a ranked list of documents as an output, a serial
examination of documents in that list with the assumption that
all documents in the list have  at least one query term. The user is expected to browse this list until the information need is satisfied and if not, at some point the user will re-formulate and re-issue the query and repeat the process, including seeing documents already viewed in a previous search.

There is no support for a searcher's
assessment of relevance, no opportunity for the searcher to indicate to the system that a document is a good, or a bad document in terms of their information need, each search operation in the sequence is independent of other searchers and thus the system is stateless with no learning, and 
no conversation or interaction between searcher and search system.

This model is used in web search, enterprise search, search through emails and social media and searches on our own computer filespace.  The takeaway message from looking at search is that search systems do not help individuals, they tune parameters to optimise for the average person, thus diluting any personalisation or  benefit to an individual person, let alone for an individual search.

An interesting use of search tools is re-finding, the phenomenon whereby a user is searching for information s/he had previously located, but cannot now find and needs to re-read or review it for the task at hand.
The concept has been around for more than a decade and one of the of the earliest studies on this was \cite{10.1145/1718487.1718512} who found that 40\% of search queries at that time, were associated with information re-finding.
Recently re-finding has morphed into what is called personalised search where stateless search systems are augmented with external resources, like contextual information  \cite{zhou2020enhancing} in order to re-find information.
We shall return to the importance of re-finding later.

\section{Computer Vision and Image Search}

Computer vision and multimedia information processing have made extreme progress within the last decade and many tasks in some domains can be performed with the same level of accuracy as if done by humans, or better. This is mostly because we  leverage the benefits of having huge amounts of data available for training, we have an enormous increase in the amount of computer processing and emergence through using GPUs in parallel, and we have seen the evolution of machine learning as a suite of techniques to combine data and processing and deliver accurate vision-based systems.
While issues of bias, explainability, compute cost and carbon footprint in training each remain as hurdles, there is now such momentum in the neural networks approach to visual information processing that these hurdles will be overcome in time.

Given that we can automatically process some forms of visual information better than humans what do we use this processing for~? We  use this  in autonomous vehicle navigation though that is a narrow and niche application as are security applications when searching CCTV for a particular individual or when classifying crowds of people for shopping and footfall analysis. Computer vision is also used for  medical image analysis for healthcare diagnostics, for remote monitoring of premises for outlier detection or monitoring of animals or crops in agriculture. All these applications and many more are widespread and established yet one which is not so widespread is image or video search directly by users. Most of the images and videos delivered to us as part of work or leisure or entertainment come from recommendations, or are included as part of a lager multimedia ``package'' like a movie, a news article, or a social media post where other aspects of that package are used to select it for delivery to us.

The types of image search which are available to us include searching for duplicates or near duplicates through reverse image search with systems like TinEye (\url{https://tineye.com/}). We also have image similarity where given an image as a search query, systems like those offered by the major web search systems, visually similar images can be found, with visual similarity based on low level features, with lots of options to filter the results. but what if we don't have a query image that we want to search against~? In such cases we are faced with having to sketch a query image, which we are terrible at doing, or we have to use words to describe what we think might be in the image we are looking for.  If we are searching for an image which we have already seen at some point in the past and we wish to re-find it, then likewise there is no support except to re-trace the route used to locate it first time, by following your browsing history for example to to try to formulate a query using words. That is because the way we index and represent images in most systems is by low level features like colours and shapes combined with tagging them with a bag of words representation and that is done at indexing time, not at search time.  This is a clear mis-match to the kind of image representation we would like to see when we try to find or re-find images. It ignores all the characteristics of human memory that lead us to need to find or to re-find the image in the first place. So no connecting an image with things we already know so it comes to mind more easily, no repeated exposure to an image so that its forgetting curve is less steep, no interaction during search or information discovery except the stateless rapid-fire query-examine-query cycle. In the final section of this paper we will look at the opportunity for a better form of image search.

\section{Image Search Should Be A Conversation}

Given the mis-match between on the one hand human memory and its characteristics of forgetting and our typical information seeking behaviours where we have tasks that require information in order to complete them, and on the other hand the typical feature representation of images, this is both a problem, and an opportunity for computer vision to do something differently.  Each image or video clip will have a natural inherent memorability associated with it, as demonstrated in the annual MediaEval video memorability benchmark evaluation ~\cite{de2020overview} and some systems based on a typical machine learning paradigm of training against manually annotated memorability scores and using an ensemble of models, show excellent performance in predicting such memorability~\cite{azcona2019predicting}.  However just because some visual imagery is more memorable does not mean we are going to actually remember it when we need to.  

Human memory will fail, naturally and when it does then this will lead to the need for re-finding images, for which we currently have to use the same form of search tools as we do for all our other kinds of searching.  We have grown accustomed to this because it has been the default search modality for decades.  Yet re-finding of imagery is much more complicated than re-finding text documents or web pages because we have this phenomenon of continuous partial attention as we work and hence we are less likely to be able to describe or verbalise a description of it for input to a search engine. Image searching is not as widespread or as common or as much used as it should be because we are tied to the search model developed for searching through text, whose characteristics are enough to deliver something that works for text, but not for images.

If we were able to start again from scratch and build image search tools based on user needs rather than convenience to existing systems then image search would be interactive, conversational and incremental rather than a rapid-fire iterations of query re-formulations.  The selection of candidate images to present to the searcher during such an interactive conversation would be based on novelty as well as similarity to any positive indications already provided by the searcher through relevance feedback. Relevance feedback would present alternative images to choose from rather than a ranked list, along the lines of ``which of these two images is closer to what you are looking for'' because rapid differentiation is something for which we are good at, and would re-enforce and clarify our actual information need.

In turn this would require computer vision to be incremental and run in real time, analysing and representing images and videos in some generic initial form and then, at query time,  re-analysing in the context of a specific query in order to determine whether the image has or has not got the features the user is looking for. That means searching would be slower to complete because of the additional query-time computation required, but more accurate and satisfactory for the user because the user would be more engaged in the search process, and once the computation of query-specific features of an image or video was done it  would be kept and re-used for subsequent querying.
Building systems such as outlined here is now within our scope and would allow image search to be more integrated and useful, within our information seeking and discovery activities.

%
%
%
\bigskip

\noindent 
{\bf Acknowledgements}:
This work is part-funded by 
Science Foundation Ireland under Grant Number SFI/12/RC/2289\_P2, co-funded by the European Regional Development Fund.

\bibliographystyle{splncs04}
\bibliography{bibliography}

\end{document}